\documentclass[runningheads]{llncs}
\usepackage{graphicx}

\usepackage{comment}
\usepackage{amsmath,amssymb} %
\usepackage{color}
\graphicspath{{Figures/}}
\usepackage[colorlinks,citecolor=green,linkcolor=red]{hyperref}

\usepackage{amsfonts}
\usepackage{booktabs}
\usepackage{subfig}

\usepackage{algorithm}
\usepackage{algorithmic}

\usepackage[width=122mm,left=12mm,paperwidth=146mm,height=193mm,top=12mm,paperheight=217mm]{geometry}
\usepackage[algo2e]{algorithm2e}

\begin{document}
\pagestyle{headings}
\mainmatter
\def\ECCVSubNumber{100}  %

\title{NullSpaceNet: Nullspace Convoluional Neural Network with Differentiable Loss Function} %

\author{Mohamed H. Abdelpakey$^{1,2}$ and Mohamed S. Shehata$^{1,2}$
\\
$^1$ University of British Columbia, \\ Department of Computer Science, Math, Physics, and Statistics.\\
Kelowna, BC, Canada\\
$^2$ Memorial University of Newfoundland,\\ Computer Engineering and Applied Science, St. John's, NL, Canada  \\
\email{\{mohamed.abdelpakey,mohamed.sami.shehata\}@ubc.ca}\\
}
\institute{Paper ID }

\maketitle

\begin{abstract}

We propose NullSpaceNet, a novel network that maps from the pixel level input to a joint-nullspace (as opposed to the traditional feature space), where the newly learned joint-nullspace features have clearer interpretation and are more separable. NullSpaceNet ensures that all inputs from the same class are collapsed into one point in this new joint-nullspace, and the different classes are collapsed into different points with high separation margins. Moreover, a novel differentiable loss function is proposed that has a closed-form solution with no free-parameters.\\
NullSpaceNet exhibits superior performance when tested against VGG16 with fully-connected layer over 4 different datasets, with accuracy gain of up to $4.55\%$, a reduction in learnable parameters from $\sim 135M to \sim19M$, and reduction in inference time of $\sim99\%$ in favor of NullSpaceNet. This means that NullSpaceNet needs less than $1\%$ of the time it takes a traditional CNN to classify a batch of images with better accuracy.\\

\end{abstract}

\section{Introduction}
In  recent years, Convolutional Neural Networks (CNNs) have revolutionized computer vision tasks such as object tracking  \cite{abdelpakey2019domainsiam,abdelpakey2019dp,mohamed2018denssiam,bhat2019learning}, surveillance systems \cite{kendall2015posenet}, image understanding \cite{lenc15understanding}, computer interactions \cite{molchanov2016online} and generative models \cite{huang2019generative}. Image classification  is one of the core  tasks in computer vision, especially in Large Scale Visual Recognition Challenges (e.g., ILSVRC15) \cite{russakovsky2015imagenet}. Most classification networks consist of two parts: 1) the feature extractor and 2) the classifier. The feature extractor uses a stack of convolutional layers to extract the deep features from the input images through consecutive convolutional operations. The classifier uses fully-connected layers  with a softmax layer. It has been proven that most of the network's learnable parameters are located in the fully connected layers \cite{krizhevsky2012imagenet}. For example, the classifier in VGG16 has $102.76$ million parameters, while the feature extractor has only $32$ million parameters. Consequently, this huge amount of learnable parameters causes a heavy load in the training phase. 

In this paper, we propose NullSpaceNet, a novel network that maps from the input pixel level to a joint-nullspace, as opposed to a traditional CNN that maps the input pixel level to a feature space. The newly learned nullspace features have a clear interpretation and are more separable. In particular, instead of using the fully connected layers with the categorical cross-entropy to maximize the likelihood between the estimated class probabilities and the true probability of the correct class, NullSpaceNet projects the pixel level inputs onto a joint-nullspace. All inputs from the same class are collapsed into one point in this new joint-nullspace and the different classes are collapsed into different points with high separation margins. Moreover, the hyperplane that  has the  orthonormal vectors of the projected nullspace features is well-defined and can be described as shown in Eq. \ref{eq:hyperplane} and Fig.  \ref{f:geometry}. In contrast to a traditional CNN used in a classification task, which optimizes its weights by maximizing the likelihood between the estimated class probability of the network's output and the true probability, NullSpaceNet minimizes the within-class scatter matrix (to be zero or very close to zero), while maintaining the between-class scatter matrix to be always positive. This makes the classification task  more robust as shown in Fig.  \ref{f:tsne_null_cce} $(a)$ and Fig.  \ref{f:tsne_null_cce} $(b)$. The results are available online  \footnote{\url{https://github.com/NullSpaceNet}}\\
To summarize, the main contributions of this paper are three-fold:
\vspace{-2mm}
\begin{enumerate}
   \item A novel Network (NullSpaceNet) that learns to  map from the input pixel level to a joint-nullspace. The formulation of NullSpaceNet ensures that the nullspace features from the same class are collapsed into a single point while the ones from different classes are collapsed into different points.\\  
 \vspace{-3mm}
   \item A differentiable  loss function is developed to train NullSpaceNet. The proposed loss function is different from the standard categorical cross-entropy. The proposed loss function ensures that the within-class scatter matrix vanishes while maintaining a positive between-class scatter matrix.
   \item The proposed NullSpaceNet has a clear interpretation, both mathematically and geometrically.  
   \end{enumerate} 
The effect of these three contributions result in  accuracy gain of up to $4.55\%$, a reduction in learnable parameters from $\sim 135M to \sim19M$, and reduction in inference time of $\sim99\%$ in favor of NullSpaceNet. This means that NullSpaceNet needs less than $1\%$ of the time it takes a traditional CNN to classify a batch of images with better accuracy over all 4 datasets we used in testing.\\
The rest of the paper is organized as follows: Related work is presented in  section \ref{relatedwork}, then section \ref{Proposed} details the proposed NullSpaceNet.  The training and inference phases are presented in section \ref{trainingandinference}. The experimental results are presented in  section \ref{Experiments}.  Finally, section \ref{Conclusions} concludes the paper.\\

\section{Related Work}\label{relatedwork}
 \vspace{-3mm}
Linear Discriminant Analysis (LDA) and nullspace have existed as analytical methods for a significant period of time \cite{zheng2017object,foster1986rank,recht2011null,kueng2017robust,eltantawy2020local,dufrenois2016null}. LDA has  been frequently been employed as a dimensionality reduction tool or feature extractor within the filed of classification \cite{guo2006null,wen2018robust,hu2019multi,hou2019linear,hou2019linear,fei2020deep,wang2017locality,korkmaz2017expert,mahdianpari2018fisher,liang2018classification}. Nullspace can be derived from the Fisher-criterion objective function in an analytical way. This work \cite{eltantawy2018krmaro} used multiple local nullspaces to detect the small moving objects in aerial videos.  Using the nullspace allows the detector to nullify the background while maintaining the moving objects.\\ 
Nullspace has been used in \cite{liu2017incremental} to specify whether  the incoming data belongs to the existing class  or not. In particular, they used Incremental Kernel Nullspaec Discriminitve (IKNDA). To speedup their method, an intelligent update scheme is used to extract information from newely added samples. The work in \cite{Tianyu2018max} proposed  Max-Mahalanobis distribution (MMD) using LDA to improve the the robustness of the adversarial attack. \cite{zhang2018cappronet} proposed to learn a capsule subspace  using orthogonal projection. The length of the resultant capsules is utilized to score the probability of belonging to different categories.
 The authors in \cite{pmlr-v77-pan17a} proposed to apply the Hybrid Orthogonal Projection Estimation (HOPE)  to CNN for image classification.  HOPE is a hybrid model that combines orthogonal linear projection, for feature extraction, with mixture models. The idea in HOPE to allow for extraction of useful information from high-dimension feature vectors while filtering out irrelevant noise. \cite{tian2017deep} used LDA with the Fisher-criterion on VGG16 to classify facial gender. LDA was applied on the output of the last layer to derive a light weight version of VGG16. A Bayesian classification is then used to classify the output. Notice that, this is completely different from NullSpaceNet, where we reformulate the learning process in a differentiable way to train the network to learn a joint-nullspace. DeepLDA \cite{Drofer2016very} proposed to  use LDA to learn to maximize  eigenvalues of the Fisher-criterion. After training, DeepLDA uses the entire training set to extract  the dominant basis vectors to project the new samples. In contrast to all previous methods, we use the nullspace in VGG16 in a learnable way with a differentiable loss function to project the pixel  level input to a joint-nullspace. \\
The only work we found that included using LDA in a deep learning framework was presented in \cite{Drofer2016very}, where the authors solved the LDA and integrated it in a deep CNN. It is worth mentioning that the work in \cite{Drofer2016very} did not include any reference to any usage of nullspace. In our work, we do not solve for the LDA, instead we reformulate the problem  within the nullspace to train the network to project from the pixel level onto the joint-nullspace.

\section{Proposed Method: NullSpaceNet}
\label{Proposed}

\subsection{Problem Definition}
\label{Problem Definition}
Given a dataset of training images $X= \{x_1,x_2,...,x_N\} \in  \Bbb{R}^{w \times h \times d}$, where $w, h$  and $d $ are the width, height, and depth of each image, respectively and $N$ is the number of images in the training dataset. Each image is associated with a respective class $C$, where $C=\{c_1,c_2,...,c_n\}  \in \Bbb {R}$, $n$ is the number of classes in the training dataset. In this paper, we use the VGG16 as the backbone network, hence, the training images will be fed into the feature extractor part $\phi(x; \theta)$. The objective is to force the network to learn a joint-nullspace that maps from the pixel level to a strong discriminative nullspace. The learned nullspace will replace the classifier part, which has the most network's learnable parameters.
\subsection{Proposed Architecture}
\label{Proposed Architecture}
\begin{figure}{!t}
 	\centering
		{\includegraphics[width=0.55\columnwidth]{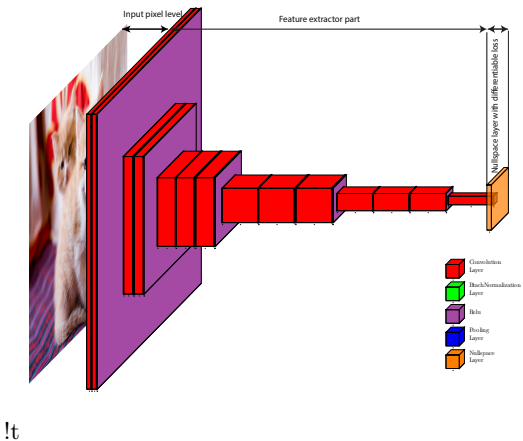} }  
	\caption{NullSpaceNet architecture, the feature extractor part is inherited from $VGG16$. The NullSpace layer is at the end of the architecture. Note that the nullspace layer has been magnified for the sake of visualization. }
	\label{f:architecture}
\end{figure}

NullSpaceNet inherits its architecture from VGG16 . Our contribution is the addition of the nullspace layer  as shown in Fig.  \ref{f:architecture} . Also, we added a (Conv-BatchNormalization-Relu) layer with kernel size=3 to produce a 2D tensor of shape $1\times800$  before the nullspace layer, in the case of STL10 dataset. In the case of CIFAR10 and CIFAR100, we change the kerenl size=1  of the last layer.   NullSpaceNet has 19 layers, each layer consists of (Conv-BatchNormalization-Relu), we consider the pooling as a stand-alone layer.\\
The novelty of NullSpaceNet lies in the nullspace layer and the diffrentiable loss function we propose in section \ref{Mathematical Formulation}. The nulllspace layer forces the network, through the  backpropagation,  to learn the projection from the input pixel level onto a joint-nullspace,  where the joint-nullspace features have optimal separation margins. The Nullspace layer achieves  this through spanning   vectors of the optimal  within-class scatter matrix as it will be discussed in more details in section \ref{Mathematical Formulation}. Formulating the nullspace layer in this way prevents the network from the Small  Sample Size (SSS) problem (i.e., the model has a high dimensional output features  while training on small batches of images).
 
\subsection{Mathematical Formulation of The Loss Function}
\label{Mathematical Formulation}
\textbf{Background:} To derive a differentiable loss function to train the joint-nullspace, we start from the linear discriminant analysis (LDA) \cite {foley1975optimal}. In this paper, we assume that the output of the feature extractor part  in the network for each image is  $F \in \Bbb{R}^{D \times 1}$, where $D$ is the depth of the output. The objective of the LDA is to find a projection matrix ${P} \in \Bbb{R}^{D \times M}$ that minimizes the within-class scatter matrix and maximizes the between-class   scatter matrix simultaneously. This can be achieved by maximizing the Fisher-discriminant criterion $\mathcal{J}({P})$ as follows:

\begin{equation}\label{e:fisher}
\mathcal{J}({P}) = \frac{{P}^{\top}{S}_{b} {P} }{{P}^{\top}{S}_{w} {P} }
\end{equation}
Where $P$ is the projection matrix, $S_b$ and $S_w$ are the between-class and within-class scatter matrices, respectively. The optimization of Eq. \ref{e:fisher} can be solved for the generalized  eigenvalue  problem as follows:

\begin{equation} 
{S}_{b} {P} = \lambda {S}_{w} {P}
\end{equation} 
where $\lambda$ is the eigenvalue of the eigenvector ${S}_{w}^{-1}{S}_{b}$. \\

\textbf{Derivation of the proposed novel  loss function:}\\
\textbf{Lemma 1:}
By investigating the output of the  network's feature extractor, it turns out that the network has a tendency to minimize the within-class scatter matrix, which is the constraint in the denominator of Eq. \ref{e:fisher}. However, it does not put constraints on the between-class  scatter matrix as shown in Fig. \ref{f:tsne_null_cce}(b).\\
\textbf{Proof:} 
The visualization in Fig.  \ref{f:tsne_null_cce} (b). The learned features of VGG16 with FC layer are scattered with no constraints on the between-class scattered matrix where some classes (e.g.,class $\#$4, $\#$5 and class $\#$2 and $\#$8) are overlapped.  
Based on Lemma 1, we force two constraints on the learning process. In particular, we force the between-class scatter matrix to always be positive while minimizing the within-class scatter matrix to be zero as follows:
\begin{equation}\label{eq:cond1} 
{P}^{\top}{S}_{b} {P} > 0
\end{equation}
\begin{equation}\label{eq:cond2} 
{P}^{\top}{S}_{w} {P} = 0
\end{equation}
\textbf{Lemma 2:}
When the network satisfies the two constraints in Lemma 1, the distribution of the same class  features in  the new joint-nullspace approaches the  Dirac Delta function.\\
\textbf{Proof:} Assuming the features are represented by the normal distribution, for simplicity, we will use a 1-D dimensional normal distribution. 

\begin{equation}\label{eq:dirac1} 
\int  g(\bar{x}) f \left(\bar{x} | \bar{\mu}, \bar{\sigma}^{2}\right) d\bar{x}
\end{equation}
where  $g(\bar{x})$ is the mean value function of the projected features by the network onto the joint-nullspace and $\bar{\sigma}$ is the standard deviation of the distribution. We take the limit of Eq. \ref{eq:dirac1} as $\bar{\sigma^2}$ approaches 0. 
\begin{equation}\label{eq:dirac2} 
\lim_{\bar{\sigma}^{2}\to 0}(\int g(\bar{x}) f(\bar{x} | \bar{\mu}, 0) d \bar{x})=\int g(\bar{x}) \delta(\bar{x}-\bar{\mu}) d \bar{x}=g(\bar{\mu})
\end{equation}
Using Lemma 1 and Lemma 2 to find the limit of Eq. \ref{e:fisher} (which guarantee the best separability as explained above), we get:
\begin{equation}\label{eq:limoffisher} 
 \lim_{(P^T S_w P)\to 0} \mathcal{J}({P}) = \infty
\end{equation}
Since the between-class scatter matrix $S_b$ in Eq. \ref{eq:cond1} is hard to calculate, especially in the case of high dimensional features, we calculate $S_b$ using the total-class scatter matrix $S_t$ and the within-class scatter matrix $S_w$ (see Eq. \ref{eq:scatters} for mathematical definition of them) as follows:
\begin{equation}\label{eq:totalscatter} 
 S_b = S_t - S_w
\end{equation}
By substituting  Eq.\ref{eq:totalscatter} in Eq.  \ref{eq:cond2} we get (since the derivation is long we will provide the details in the supplementary material):
\begin{equation}\label{eq:modified_constarint} 
{P}^{\top}{S}_{t} {P} > 0
\end{equation}
Since the output of the NullSpaceNet is  $f \in \Bbb{R}^{D \times N}$ when the input batch images $X \in \Bbb{R}^ {w \times h \times d \times N}$ , where $N$ is the number of images. We define the the within-class scatter matrix $S_w$ and the total-class scatter matrix  $S_t$ from the output of NullSpaceNet as follows:
\begin{equation}\label{eq:scatters}
{S}_{w} = \frac{1}{N} {F}_{w}^{\top} {F}_{w},\quad {S}_{t} = \frac{1}{N} {F}_{t}^{\top} {F}_{t}
\end{equation} 
where $F_w$ is the centered class mean output features (i.e, subtracting the class mean from each feature output belonging to this class), and $F_t$ is the centered global mean output features as shown in Eq. \ref{eq:means}.\\

\begin{equation}\label{eq:means}
{F}_{w} = \frac{1}{N} {(X-\mu_{c})}^{\top} ,\quad {F}_{t} = \frac{1}{N} {(X-\mu_{g})}^{\top} 
\end{equation}
Where $\mu_{c}$ is the class mean and $\mu_{g}$is the global mean of the dataset.\\
Now, we want to integrate the scatter matrices the we derived in Eq. \ref{eq:scatters} in the joint-nullspace formulation. Let $U_t$ denote the nullspace of the total-class scatter matrix and $U_w$ denote the  nullspace of within-class matrix. From the definition of the nullspace and using the fact that $S_t$ is non-negative definite, we get:
\begin{equation}\label{eq:nullspace_ut}
\begin{split} 
{U}_{t} &=  \left \{  {u}\in \mathbb{R}^{D}\mid {S}_{t}{u}=0\right \}  = \left \{{u}\in \mathbb{R}^{D}\mid   {u}^{\top}{S}_{t}{u} = 0 \right \} \\
&= \left \{{u}\in \mathbb{R}^{D}\mid   ({F}_{t}{u})^{\top}{F}_{t}{u} = 0 \right \} \\
&=  \left \{  {u}\in \mathbb{R}^{D}\mid {F}_{t}{u}=0\right \}.
\end{split}
\end{equation}
similarly, we get $U_w$ (details are provided in the supplementary material).\\

\textbf{Lemma 3:} The projection matrix $P$ that satisfies the constraints in Eq. \ref{eq:cond1} and Eq. \ref{eq:cond2} can be achieved, if and only if,  $P$ lies in the shared space between $U_t ^ \perp$ and $U_w$ as shown in Eq. \ref{eq:overlabspacese}. 
\begin{equation} \label{eq:overlabspacese}
{P} \in ({U}_{t}^{\perp}\cap  {U}_{w}).
\end{equation} 
 where $U_t ^ \perp$ is the orthogonal complement subspace of $U_t$  spanned by the  the centered global mean output features, it can be obtained using  the Gram-Schmidt process \cite{jensen1977mathematical}.\\
\textbf{Proof:} Geometrically by looking at  Eq. \ref{eq:nullspace_ut} and $U_w$, the only space that satisfies  $S_t u=0$ and $S_w u= 0$ is the joint-space where $U_t^\perp$ and $U_w$ are overlapped \cite{guo2006null}.\\
Now we have the nullspace of $S_w$  which is $U_w$ and the nullspace of $S_t$ which is $U_t$. One problem with the calculation the nullspace of $S_w$ is that the dimensionality of nullspace is at least $(D+C-n)$, where $D$ is data dimensionality (which is high when we use the output of NullSpaceNet, e.g., $2048$), $C$ is the number of classes, and n is the sample sizes as it has been proved in \cite{chen2000new}. To address this problem, we revert to Eq. \ref{eq:totalscatter} where it can be seen that $S_t$ is the intersection of the nullspace of $S_b$ and the nullspace of $S_w$. Hence, the nullspace of $S_t$ can be removed based on this observation. We proceed with the solution using the Singular Value Decomposition (SVD) theory to decompose $F_t$ (which was introduced in Eq. \ref{eq:means}) as follows: 
\begin{equation} \label{eq:svd}
F_{t}=U \Sigma V^{T}
\end{equation} 
Where $U$ and $V$ are orthogonal. 

\begin{equation} \label{eq:sig}
\Sigma=\left(\begin{array}{cc}
\Sigma_{t} & 0 \\
0 & 0
\end{array}\right)
\end{equation} 
 $\Sigma_t$ is the diagonal matrix $\Sigma_{t} \in \mathbb{R}^{t \times t}$ with the eigenvalues. Now we can represent $S_t$ as follows:

\begin{equation} \label{eq:S_t}
\begin{split} 
S_{t} &=   F_{t} F_{t}^{T} \\ 
&=  V \Sigma^{T} U^{T} U \Sigma V^{T} \\
&= U \Sigma \Sigma^{T} U^{T} \\ 
&= U\left(\begin{array}{cc} 
\Sigma_{t}^{2} & 0 \\
0 & 0
\end{array}\right) U^{T}
\end{split}
\end{equation}
We select a part of orthogonal basis of $U$  with dimension $U1 \in \mathbb{R}^{m \times t}$ where $t =RANK (S_t)$, using the new subspace $U_1$ spanned by the new set of the orthogonal basis we project the scatter matrices as follows: 
\begin{equation}\label{eq:projected_scatters}
\tilde{S}_{b}=U_{1}S_{b} U_{1}^{T} , \tilde{S}_{w}=U_{1} S_{w} U_{1}^{T}, \text { and } \tilde{S}_{t}=U_{1} S_{t} U_{1}^{T}
\end{equation}
Where $\tilde{(.)}$ represents the reduced version of the decomposed $S_b, S_w, and S_t$.\\
From Eq. \ref{eq:svd} and Eq. \ref{eq:projected_scatters} we can apply the SVD again on $F_t$ with complexity of $O\left(D n^{2}\right)$ instead of $O\left(D^{2} n\right)$. Now 
instead of calculating the nullspace of  $S_w$, we can calculate the nullspace of $\tilde{S}_{w}$ as shown in Eq. \ref{eq:nullspace_w}. This gives the  network two advantages: 1) the model does not suffer from the small sample size (SSS) problem, e.g., the model has a high dimensional output features  while training on small batches of images,  as in \cite{Drofer2016very}, and 2) it is faster than solving the generalized eigenvalue problem.

\begin{equation} \label{eq:nullspace_w}
W = Span( \tilde{S}_{w})
\end{equation}
Where $W$ is the nullspace of $\tilde{S}_{w}$. Finally, the projection matrix that satisfies Eq. \ref{eq:cond1} and Eq. \ref{eq:cond2} can be calculated by: 
\begin{equation} \label{eq:projection}
P = W \times M
\end{equation}
Where M is the  eigenvectors of $W^{T} \tilde{S}_{b} W$ corresponding to the non-zero eigenvalues.
\begin{algorithm}
	\SetKwData{Left}{left}\SetKwData{This}{this}\SetKwData{Up}{up}
	\SetKwFunction{Union}{Union}\SetKwFunction{FindCompress}{FindCompress}
	\SetKwInOut{Input}{input}\SetKwInOut{Output}{output}
	
	\Input{The output of the last layer of $VGG16$  $F \in \Bbb{R}^{D \times N}$ }
	\Output{ Optimize the weights of the NullSpaceNet  using the proposed differentiable  loss function $\mathcal{L}$  based on the nullspace formulation  } \BlankLine
	 \begin{algorithmic}[1]
	\STATE Calculate matrix $F_t$;
	\STATE Calculate $SVD (F_t^T) $;
	\STATE  Calculate the scatter matrices $S_w, S_b, S_t$;
    \STATE Calculate matrices $\bar{S_t}, \bar{S_b}, \bar{S_w}$ from Eq. \ref{eq:projected_scatters};
    \STATE  Calculate the nullspace $W$ of $\bar{S_w}$ using Eq. \ref{eq:nullspace_w};
    \STATE Solve for the eigenvlaues  of $ W^{T} \tilde{S}_{b} W $;
    \STATE Formulate  the loss function over the average of the non-zero  eigenvalues  from the last step using Eq. \ref{eq:cross_entropy};  %
    \STATE Use the proposed differentiable loss function in Eqs. \ref{eq:cross_entropy} and \ref{eq:diff_loss_func} to train the network 
	\caption{Steps to Calculate the Proposed Loss Function } \label{alg:loss_algorithm}
	\end{algorithmic}
\end{algorithm} \DecMargin{1em} 
\vspace*{-10mm}
\subsection{Gradient of the Loss Function}
Training the NullSpaceNet requires the loss function to be differentiable everywhere. Hence, we propose a novel differentiable loss function that maximizes the positive (or minimizes the negative) of the average non-zero eigenvalues of the decomposed  $W^{T} \tilde{S}_{b} W$. We define $k$ as the number of eigenvalues in two cases: 1) when $\{E_1,..., E_{c-1} \} > 0$, and 2) when $\{E_k,..., E_{c-1} \}< 0$. The steps to calculate the proposed differentiable loss function is shown in Alg. \ref{alg:loss_algorithm}. The final equation of these steps is shown in Eq. \ref{eq:cross_entropy}.
\begin{equation}  \label{eq:cross_entropy}
	\mathcal{L}(\phi(E; \theta)) = -\sum_{i=1}^{k} {SVD}_{E_i}(W^{T} \tilde{S}_{b} W)
\end{equation}
 Where  $E\in$ $\left\{E_{1}, \ldots, E_{k}\right\}=\left\{E_{j} | E_{j}<\min \left\{E_{1}, \ldots, E_{C-1}\right\}+\epsilon\right\} $.\\
Using the chain rule, the derivative of the loss function in Eq. \ref{eq:cross_entropy} w.r.t the last  layer $H$ of NullSpaceNet is given by Eq. \ref{eq:diff_loss_func} (details are given  in the supplementary materials).
\begin{equation} \label{eq:diff_loss_func}
\begin{split} 
&\frac{\partial \mathcal{L}}{\partial {H}} 
= -\sum_{i=1}^{k} \frac{\partial  {SVD}_{E_i}(W^{T} \tilde{S}_{b} W)}{\sum_{i=1}^{k} {\psi}_{i}^{T}\left(\frac{\partial {\tilde{S_b}}}{\partial {H}}-\psi_{i} \frac{\partial {\tilde{S_w}}}{\partial {H}}\right) {\psi}_{i}}  \times  \\  
 &\frac{ \sum_{i=1}^{k} {\psi}_{i}^{T}\left(\frac{\partial {\tilde{S_b}}}{\partial {H}}-E_{i} \frac{\partial {\tilde{S_w}}}{\partial {H}}\right) {\psi}_{i}}{\partial {H}} 
\end{split}
\end{equation}
Where $\psi_i$ are the Eigenvectors  associated with the Eigenvalues $E_i$

\subsection{Insights into NullSpaceNet}
\label{interpretation}
\begin{figure}
 	\centering
		{\includegraphics[width=0.65\columnwidth]{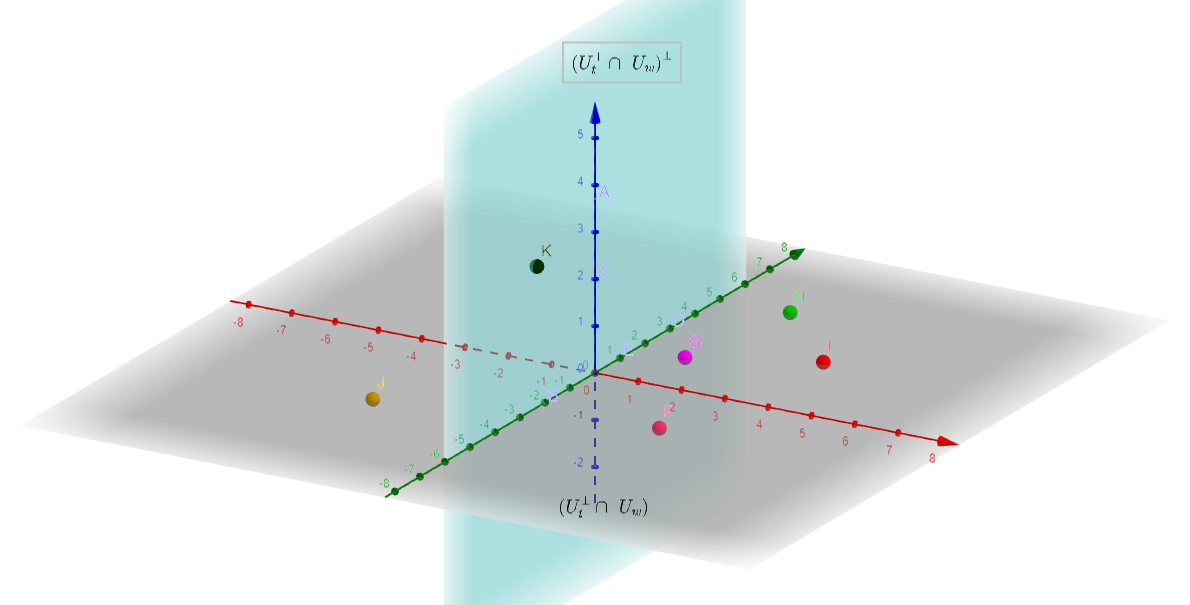} }  

	\caption{Geometric illustration of the  learned feature projected by the network in the joint-nullspace. Each color associated with a letter encodes a class, note that all inputs from the same class are collapsed into a single point. Notice that all classes exist on the Grey-colored hyperplane ($U_t  ^\perp \cap U_w$) including point $k$. }
	\label{f:geometry}
\end{figure}

In this section, we provide a deeper look, both mathematically and geometrically, into the proposed NullSpaceNet.\\
\textbf{Mathematical Insights: }
The main idea of NullSaceNet is to learn to project the input data onto another subspace (different from the traditional feature space) that satisfies the two constraints in Eq. \ref{eq:cond1} and Eq. \ref{eq:cond2}. The new proposed subspace (i.e., joint-nullspace) mathematically forces the within-class scatter matrix to vanish through the optimization of the proposed loss function in Eqs. \ref{eq:cross_entropy} and \ref{eq:diff_loss_func}. Meanwhile the new joint-nullspace mathematically forces the between-class scatter matrix to always be positive through  the optimization  of the loss function in Eq. \ref{eq:cross_entropy}.\\ 
\textbf{Geometric Insights: }
The features that are produced by NullSpaceNet are living in the hyperplane represented  by $U_t  ^\perp \cap U_w$, and all inputs from the same class are collapsed into one point, while the inputs from different classes are collapsed into different points as shown in Fig.  \ref{f:geometry}. The hyperplane now is well-defined and all the features are located in a confined space that can be precisely described both mathematically and geometrically. 

\vspace*{-10mm}
\section{Training and Inference of NullSpaceNet}
\label{trainingandinference}
\subsection{ NullSpaceNet Training Phase}
Firstly, the input batch of images is fed into the input layer as shown in Fig. . \ref{f:architecture}. The batch undergoes the operation of the layers such as convolution, normalization, and pooling, then to the nullspace layer, where all calculations and the loss function are performed as shown in Alg. \ref{alg:loss_algorithm}.
\subsection{  NullSpaceNet Inference Phase}
Even though the training phase does not require a special setup, the inference phase in the NullSpaceNet is different. After the network is trained using the proposed differentiable loss function in Eq. \ref{eq:cross_entropy}, we feed the entire training set into NullSpaceNet, then  extract the output mean of each class from the last layer $\mu_{k}=(\mu_{1},...,\mu_{C})$ that has the dimension $\mathcal{R}^{D\times N}$, where $N$ is the number of images. The eigenvectors  of the decomposed   $W^{T} \tilde{S}_{b} W$ will be calculated, the number of the basis vectors  are $C-1$, where $C$ is the number of training classes. From Eq. \ref{eq:projection}, the projection matrix $P$ is calculated. Using the hyperplane equation, any output of the testing dataset $t^T$ can be classified using the hyperplane equation \cite{friedman2001elements} (Eq. \ref{eq:hyperplane}) below: 
\begin{equation}\label{eq:hyperplane}
\operatorname{argmax}_{k} \beta_{k}(t)=t^{T} \mathbf{\Sigma}^{-1} \mu_{k}-\frac{1}{2} \mu_{k}^{T} \mathbf{\Sigma}^{-1} \mu_{k}
\end{equation}
Where $\Sigma = P\times P^T$ and $\beta$ is the hyperplane.

\section{Experimental Results}
\label{Experiments}
\textbf{Implementation Details}\\
NullSpaceNet  has been implemented in Python using PyTorch framework \cite{paszke2017pytorch}. All experiments have been  performed on  Linux with Xeon E5 $@$2.20 GHz CPU and  NVIDIA Titan XP GPU. All experiments are performed on networks trained from scratch. We set $\epsilon$ to 1 and the number of epochs for the training to 200. We used Adam optimizer \cite{kingma2014adam} with learning rate of 0.0001, a momentum of 0.9, and  a batch size  of 400 images. \\
\textbf{Datasets}\\
 The datasets CIFAR10 and CIFAR100 \cite{krizhevsky2009learning} have resolution of  32 $\times$ 32. CIFAR10  has 10 classes collected from natural images, while CIFAR100 has 100 classes. Each dataset has 50,000 images for training and 10,000 images for testing. We used 49,000 for training, 1,000 for validation and 10,000 for testing for both datasets.  STL10 dataset \cite{coates2011analysis} has 10 classes with higher image resolutions, in contrast to CIFAR datasets, to show the effect of higher image resolution on the performance. STL10 has 5,000 images for training, while the testing set has 8,000 images. We used 5,000 for training set, 1,000 for validation set from the testing set and the remaining 7,000 for testing.\\
\begin{figure} [!t]
 	\centering
		\subfloat[NullSpaceNet] {{\includegraphics[width=0.4\columnwidth]{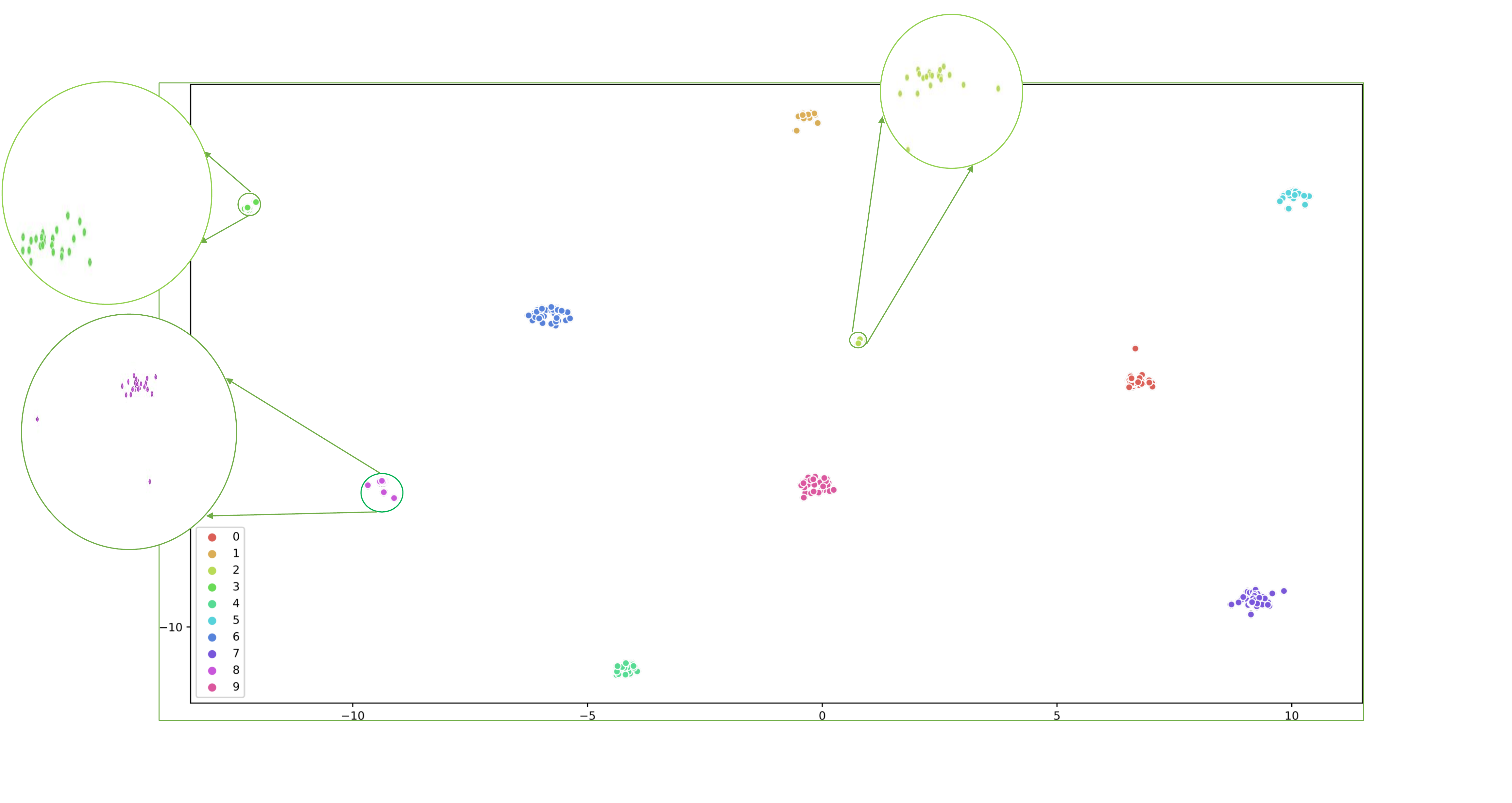} }}  
		\qquad
		\subfloat[VGG16 with FC] {{\includegraphics[width=0.4\columnwidth]{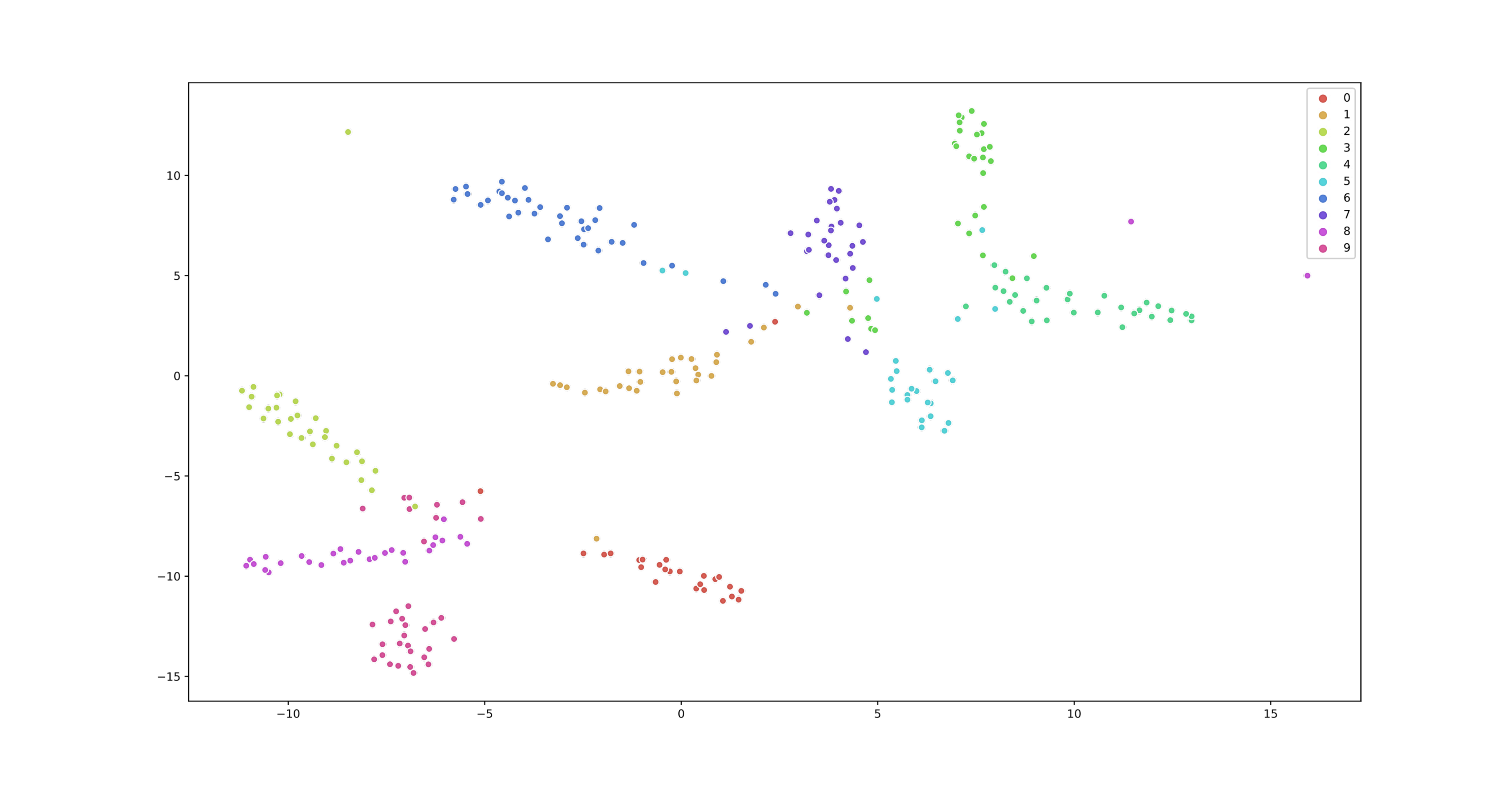} }} 
	\caption{t-SNE visualization of the learned features on STL10 dataset using (a) NullSpaceNet, (b) VGG16 with FC layer.}
	\label{f:tsne_null_cce}
\vspace{-5 mm}
\end{figure} 
\textbf{Results on CIFAR10 Dataset}\\
The results on CIFAR10 dataset are shown in Table \ref{t:Cifar10_not_pretrained}.
VGG16 with fully-connected (FC) layer and categorical-cross entropy achieves  a test accuracy of $93.51\%$, while the proposed NullSpaceNet achives $94.01\%$.\\
The accuracy difference between the proposed NullSpaceNet  and the VGG16 with FC layer is $\sim 0.5\%$, in favor of NullSpaceNet. More importantly, there is a significant reduction in the network parameters of NullSpaceNet compared to VGG16 with FC layer. The parameters went down from $\sim 134$ million in VGG16 with FC layer to $\sim 18$ million in NullSpaceNet, which is a reduction of $92.99\%$. Table \ref{t:inference_time_CIFAR} shows the inference time required per batch (2000 images), the time required by VGG16 with FC is $0.6841$ seconds while NullSpaceNet required only $0.0051$ secconds, which is a reduction of $99.25\%$ in favor of NullSpaceNet.\\
\textbf{Results on CIFAR100 Dataset}\\   
The results on CIFAR100 dataset are shown in Table \ref{t:Cifar100_not_pretrained}.
 NullSpaceNet outperforms VGG16 with FC layer by $0.07\%$; still the gain is not significant similar to the CIFAR10 dataset. But, the number of parameters in NullSpaceNet has been reduced from $\sim 134$ millions to $\sim 18$ million parameters. Table \ref{t:inference_time_CIFAR} shows the inference time required per batch (2000 images), the time required by VGG16 with FC is $0.6841$ seconds while NullSpaceNet required only $0.0051$ secconds, which is a reduction of $99.25\%$ in favor of NullSpaceNet.\\
The importance of conducting this experiment on CIFAR100 dataset is to prove that NullSpaceNet performance is not affected by the increase in the number of classes in the classification task.\\ 
\textbf{Results on STL10 Dataset}\\ 
The results on STL10 dataset are shown in Tables \ref{t:STL10_not_pretrained} and \ref{t:inference_time_STL}. 
NullSpaceNet outperforms VGG16 with FC layer in terms of accuracy with gain of $2.57\%$, parameters reduction of $92.99\%$, and inference time reduction of $99.22\%$. It is worth noting that NullSpaceNet significantly benefits from the higher image resolution, STL10 has images resolution of 64 $\times$ 64.\\ 
To visualize the learned features by NullSpaceNet and VGG16 with FC layer on STL10 dataset, t-SNE is used to provide Fig.  \ref{f:tsne_null_cce}. Each color is associated with a number that represents a class in the STL10 dataset. It can be seen from Fig.  \ref{f:tsne_null_cce}(a) that the within-class scatter matrix for all classes has been reduced to minimum and the between-class scatter maximized the margin separation between class. By examining Fig.  \ref{f:tsne_null_cce}(b), the classes are overlapping and the separation margin is not optimal. Fig.  \ref{f:tsne_null_cce} visualizes the power of the new proposed NullSpaceNet.\\ 
\textbf{Results on ImageNet}\\ 
Results on ImageNet are shown in Table \ref{t:ILSVRC15}. Note that we used only  50,000 images from ImageNet. ImageNet has a higher resolution compared to CIFAR datasets and STL10, we have downscaled all images to be $128 \times 128$. The number of parameters have increased from $\sim$ 134 to $\sim$ 135 in VGG16+FC, while it went up from $\sim $ 18 to $\sim$ 19 in the case of NullSpaceNet. AS it can be seen from Table \ref{t:ILSVRC15}, the accuracy gain is more evident in favor of NullSpaceNet, as the resolution increases. NullSpaceNet has an accuracy of $98.87\%$ compared to $94.32\%$ in VGG16+FC, which is a gain of $+4.55\%$. The inference time for VGG16+FC is $1.5653$ seconds per batch, while NullSpaceNet needed only $0.0168$ seconds, which is a reduction of $98.92\%$.\\
\textbf{Effect of Image Resolution}\\ 
The accuracy gain between the proposed NullSpaceNet  and the VGG16 with FC layer when tested on CIFAR10 and CIFAR100 is $0.5\%$ and $0.07\%$, respectively, in favor of NullSpaceNet. The gain suggests that the accuracy does not significantly benefit from the projection onto the proposed joint-nullspace in this case. This can be justified based on the fact that the images resolution in CIFAR10 and CIFAR100 is 32 $\times$ 32. This means that the number of pixel level features to be mapped in either the feature space or the joint-null space is small, and hence explains the small accuracy gain.\\ 
This justification is further supported in lights of the results on the STL10 dataset (which has higher images resolution $64 \times 64$), and consequently better accuracy in favor of NullSpaceNet. Furthermore, another experiment has been performed on a reduced resolution version of STL10 dataset. All training images have been reduced to 32 $\times $32 resolution, similar to CIFAR10 and CIFAR100 dataset. NullSpaceNet has been trained on the modified version STL10, and the results are shown in Table \ref{t:modifed_STL10}. It is seen that the accuracy gain is similar to the ones in CIFAR10 and CIFAR100. 
Another experiment has been conducted on ImageNet, the results are shown in table \ref{t:ILSVRC15} and it shows a cccuracy difference of $4.55\%$.
This confirms our justification that NullSpaceNet power becomes more clear in cases of images with higher resolution. In general, NullSpaceNet will outperform VGG16 with FC layer in all cases. All results are summarized in table \ref{t:summerize} \\
\begin{figure}[!t]
 	\centering
		{\includegraphics[width=0.6\columnwidth]{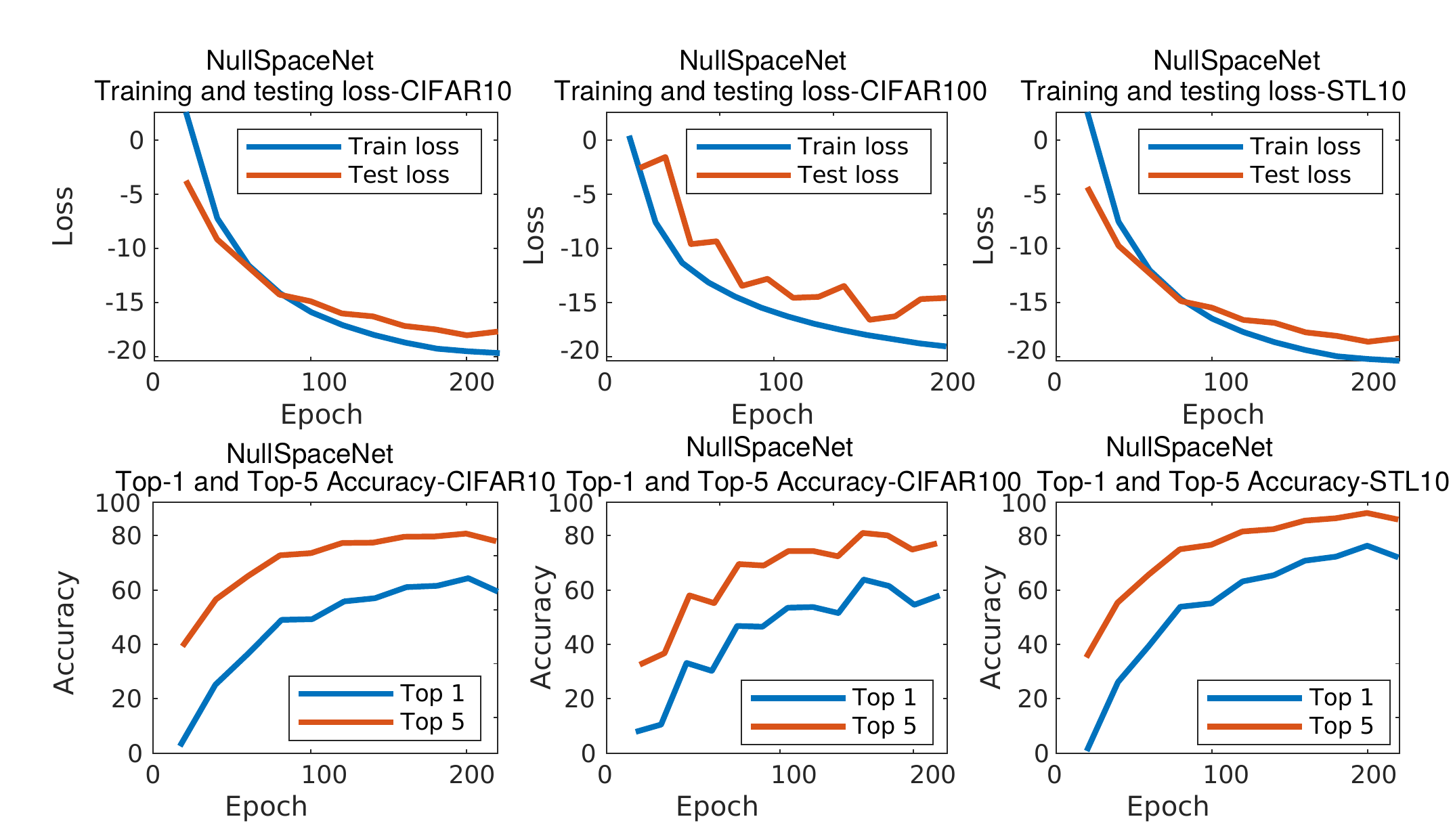} }  
	\caption{Acuuracies and Losses on CIFAR10, CIFAR100, and STL10 - Top row: Training and testing losses, Bottom row: the top-1 and top-5 accuracy. }
	\label{f:top_acc}
\vspace{-5mm}
\end{figure} 
\begin{figure}[!t]
 	\centering
		{\includegraphics[width=0.26\columnwidth]{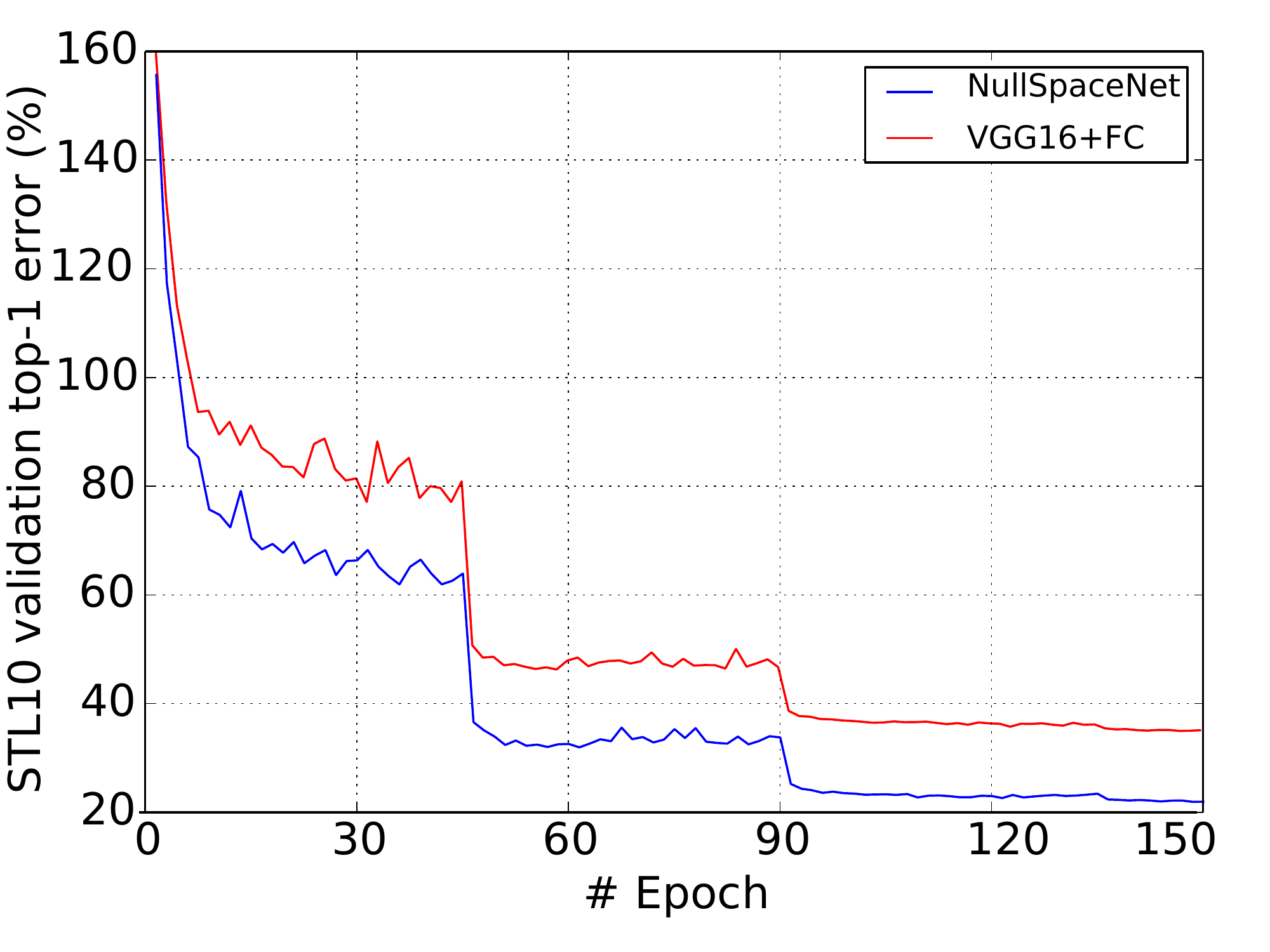} }  
	\caption{Top-1 error on the validation set of STL10. At the beginning, NullSpaceNet and VGG16+FC have almost the same error rate, then NullSpaceNet has lower error rate as the number of epochs increase. }
	\label{f:top_err}
\vspace{-5mm}
\end{figure} 
\textbf{Top-k Error Rate on STL10}\\
Top-k error rate is the fraction of the testing set for which the true label is not among the five labels that are most likely by the model prediction \cite{krizhevsky2012imagenet}. The top row shows the training and testing loss. In the bottom row, we report the top-1 and top-5 accuracy on the STL10 dataset.\\
\\

\section{Conclusions and Future Work} 
\label{Conclusions}
A typical CNN optimizes the weights of the network by maximizing the likelihood  between the estimated probability of the predicted class and the true probability of the correct class. In contrast, NullSpaceNet learns to project the features from the pixel level (i.e, input image ) onto a joint-nullspace. All features from the same class are collapsed into a single point in the learned joint-nullspace, whereas all features from different classes are collapsed into different points with high separation margins. Also, a novel differentiable loss function is developed to train NullSpaceNet to learn to project the features onto the joint-nullspace. NullSpaceNet with the proposed differentiable loss function exhibits a superior performance, with accuracy gain of $0.07-2.57\%$, and reduction in inference time of $99.22-99.25\%$ in favor of NullSpaceNet. This means that NullSpaceNet needs less than $1\%$ of the time it takes a traditional CNN to classify a batch of images with slightly better accuracy, Future work include extending this work to other fields such as object tracking. \\ 

\begin{table*}[!t] %
\centering
\caption{Test Accuracy on CIFAR10.}\label{t:Cifar10_not_pretrained}
\begin{tabular}{|l|c|r|}
\specialrule{1.2pt}{0pt}{0pt}
Architecture &  Accuracy & $\#$ Parameters\\
\hline
$VGG16$ + FC (cross-entropy loss)& 93.51$\%$ &134,309,962 \\
\hline
\textbf{NullSpaceNet (Proposed)} &\textbf{94.01$\%$}& \textbf{18,411,936} \\ 
\specialrule{1.2pt}{0pt}{0pt}
\end{tabular}
\end{table*}
\begin{table*}[!t] %
\centering
\caption{Test Accuracy on CIFAR100.}\label{t:Cifar100_not_pretrained}
\begin{tabular}{|l|c|r|}
\specialrule{1.2pt}{0pt}{0pt}
Architecture &  Accuracy & $\#$ Parameters\\
\hline
$VGG16$ + FC (cross-entropy loss)& 92.26$\%$&134,309,962\\
\hline
\textbf{NullSpaceNet( proposed )} &\textbf{92.33$\%$}& \textbf{18,411,936}\\ 
\specialrule{1.2pt}{0pt}{0pt}
\end{tabular}
\end{table*}
\begin{table*}[!t] %
\centering
\caption{Inference time and number of parameters on CIFAR10/CIFAR100}\label{t:inference_time_CIFAR}
\begin{tabular}{|l|c|l|}
\specialrule{1.2pt}{0pt}{0pt}
Architecture & $\#$ Parameters &  Average inference time/batch\\
\hline
$VGG16$ + FC (cross-entropy loss)& 134,309,962  &  0.6841 Seconds\\
\hline
\textbf{NullSpaceNet(proposed)} &\textbf{18,411,936 }&  \textbf{0.0051} Seconds\\ 
\specialrule{1.2pt}{0pt}{0pt}
\end{tabular}
\end{table*}
\begin{table*}[!t] %
\centering
\caption{Test Accuracy STL10 dataset. }\label{t:STL10_not_pretrained}
\begin{tabular}{|l|c|r|}
\specialrule{1.2pt}{0pt}{0pt}
Architecture &  Accuracy & $\#$ Parameters\\
\hline
$VGG16$ + FC (cross-entropy loss)& 93.74$\%$ &134,309,962 \\
\hline
\textbf{NullSpaceNet (proposed)} &\textbf{96.31$\%$}& \textbf{18,411,936} \\ 
\specialrule{1.2pt}{0pt}{0pt}
\end{tabular}
\end{table*}
\begin{table*}[!t] %
\centering
\caption{Inference time and number of parameters on STL10. }\label{t:inference_time_STL}
\begin{tabular}{|l|c|l|}
\specialrule{1.2pt}{0pt}{0pt}
Architecture & $\#$ Parameters & Average inference time/batch\\
\hline
$VGG16$ + FC (cross-entropy loss)& 134,309,962  & 1.3487  Seconds\\
\hline
\textbf{NullSpaceNet(proposed)} &\textbf{18,411,936 }& \textbf{0.0105} Seconds\\ 
\specialrule{1.2pt}{0pt}{0pt}
\end{tabular}
\end{table*}
\begin{table*}[!t] %
\centering
\caption{Test Accuracy on modified version of STL10 (32 $\times$ 32).}\label{t:modifed_STL10}
\begin{tabular}{|l|c|r|}
\specialrule{1.2pt}{0pt}{0pt}
Architecture &  Accuracy & $\#$ Parameters\\
\hline
$VGG16$ + FC (cross-entropy loss)& 93.89$\%$ &134,309,962  \\
\hline
\textbf{NullSpaceNet (Proposed)} &\textbf{93.91}$\%$& \textbf{18,411,936 } \\ 
\specialrule{1.2pt}{0pt}{0pt}
\end{tabular}
\end{table*}
\begin{table*}[!t]
\centering
\caption{Test Accuracy on ImageNet dataset  with image resolution $128 \times 128$. Only 50,000 images were used.}\label{t:ILSVRC15}
\begin{tabular}{|l|c|r|l|}
\specialrule{1.2pt}{0pt}{0pt}
Architecture &  Accuracy & $\#$ Parameters & Average Inference time/batch\\
\hline
$VGG16$ + FC (cross-entropy loss)& 94.32$\%$ &135,310,123 & 1.5653 \\
\hline
\textbf{NullSpaceNet (Proposed)} &\textbf{98.87}$\%$& \textbf{19,415,654 }  & 0.0168\\ 
\specialrule{1.2pt}{0pt}{0pt}
\end{tabular}
\end{table*}

\begin{table*}[!t]
\centering
\caption{A summerization of the results on 4 datasets, CIFAR10, CIFAR100, STL10, and ImagNet. All results are in favor of NullSpaceNet}\label{t:summerize}
\begin{tabular}{|c|c|c|c|c|}
\specialrule{1.2pt}{0pt}{0pt}
Dataset &  Accuracy Difference & Image Size & Parameters Reduction & Time Reduction\\
\hline
CIFAR10  &  $+0.5 \%$ & $32\times32$ & $+86.29\%$ & $+99.25\%$ \\
\hline
CIFAR100  &  $+0.07 \%$  & $32\times32$ & $+86.29\%$ & $+99.25\%$ \\
\hline
STL10  &  $+2.57 \%$ & $64\times64$ & $+86.29\%$ & $+99.22\%$ \\
\hline
STL10 (modified)  &  $+ 0.02\%$ & $32\times32$ & $+86.29\%$ & $+99.22\%$ \\
\hline
ImageNet  &  $+4.55 \%$ & $128\times128$ & +85.65\% & $+98.92\%$ \\

\specialrule{1.2pt}{0pt}{0pt}
\end{tabular}
\end{table*}

\clearpage
\bibliographystyle{splncs04}
\bibliography{egbib}
\end{document}